\documentclass[final]{cvpr}

\usepackage{times}
\usepackage{epsfig}
\usepackage{graphicx}
\usepackage{amsmath}
\usepackage{amssymb}
\usepackage{multirow}
\usepackage{verbatim}
\usepackage{subfigure}
\usepackage{booktabs}
\usepackage{makecell}
\usepackage{tablefootnote}
\usepackage{lipsum}
\usepackage{multicol}
\usepackage{caption}

\usepackage[linesnumbered,vlined,ruled]{algorithm2e}

\usepackage[pagebackref=true,breaklinks=true,colorlinks,bookmarks=false]{hyperref}
\usepackage{subfigure}
\usepackage{pifont}
\newcommand{\cmark}{\ding{51}}%
\newcommand{\xmark}{\ding{55}}%

\newcommand\blfootnote[1]{%
\begingroup
\renewcommand\thefootnote{}\footnote{#1}%
\addtocounter{footnote}{-1}%
\endgroup
}

\def\cvprPaperID{3648} 
\def\confYear{CVPR 2021}
\def\Ours{{Scale-aware AutoAug}\xspace}

\begin{document}

\title{Scale-aware Automatic Augmentation for Object Detection}

\author{
Yukang Chen$^{1}$\thanks{This work was done during an internship
at ByteDance AI Lab. Tao Kong is responsible for correspondence.
$^{\dagger}$Equal contribution.}$\,\;^{\dagger}$,
~~~
Yanwei Li$^{1\dagger}$,~~~
Tao Kong$^{2}$,~~~
Lu Qi$^1$,~~~
Ruihang Chu$^{1*}$,~~~
Lei Li$^2$,~~~
Jiaya Jia$^{1.3}$
\\[0.2cm]
$^1$The Chinese University of Hong Kong~~
$^2$ByteDance AI Lab~~
$^3$SmartMore
}

\maketitle
\pagestyle{empty}
\thispagestyle{empty}

\begin{abstract}
   We propose Scale-aware AutoAug to learn data augmentation policies for object detection.
   We define a new scale-aware search space, where both image- and box-level augmentations are designed for maintaining scale invariance.
   Upon this search space, we propose a new search metric, termed Pareto Scale Balance, to facilitate search with high efficiency. In experiments, Scale-aware AutoAug yields significant and consistent improvement on various object detectors ({\em e.g.}, RetinaNet, Faster R-CNN, Mask R-CNN, and FCOS), even compared with strong multi-scale training baselines.
   Our searched augmentation policies are transferable to other datasets and box-level tasks beyond object detection ({\em e.g.}, instance segmentation and keypoint estimation) to improve performance. The search cost is much less than previous automated augmentation approaches for object detection. It is notable that our searched policies have meaningful patterns, which intuitively provide valuable insight for human data augmentation design. Code and models will be available at \href{https://github.com/Jia-Research-Lab/SA-AutoAug}{https://github.com/Jia-Research-Lab/SA-AutoAug}.
\end{abstract}

\section{Introduction}

Object detection, aiming to locate as well as classify various objects, is one of the core tasks in the computer vision.
Due to the large scale variance of 
objects in real-world scenarios, 
it raises concerns on how to bring the scale adaptation to the network efficiently.
Previous work handles this challenge mainly from two aspects, namely {\em network architecture} and {\em data augmentation}.
To make the network scale invariant, 
in-network feature pyramids~\cite{fpn,xu2019auto,hypernet}
and adaptive receptive fields~\cite{li2019scale} are usually employed.
Another crucial technique to enable scale invariance is data augmentation, which is independent of specific architectures, and can be generalized among multiple tasks.
\begin{figure}[t]
\label{fig:comparison}
\begin{center}
   \includegraphics[width=0.9\linewidth]{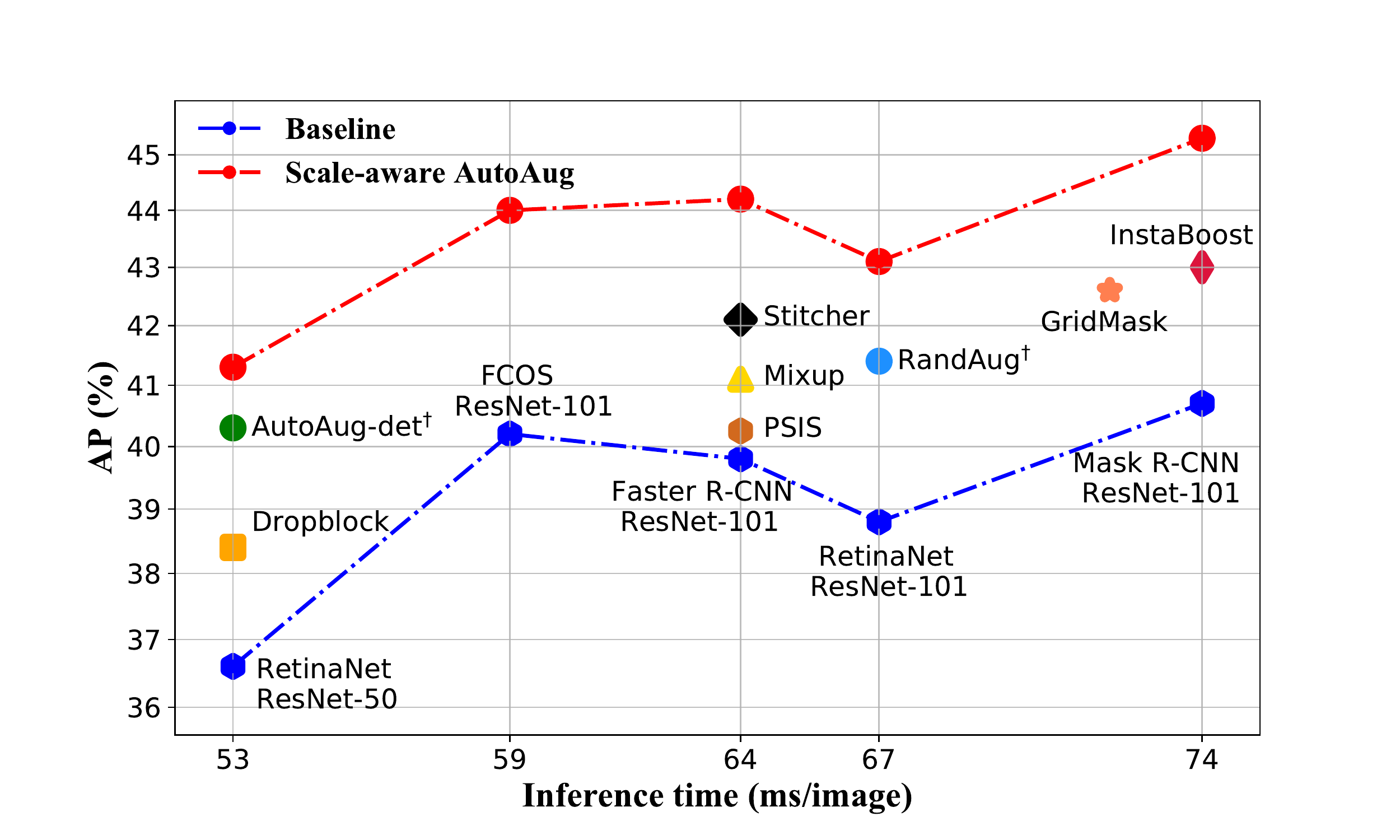}
   \caption{\small Comparison with object detection augmentation strategies on MS COCO dataset. 
   Methods in the same vertical line are based upon the same detector. Scale-aware AutoAug outperforms both hand-crafted and learned strategies on various detectors.}
\end{center}
\vspace{-1.5em}
\end{figure}

This paper focuses on data augmentation for object detection.
Current data augmentation strategies can be grouped into
color operations ({\em e.g.}, brightness,  contrast,  and  whitening) 
and geometric operations ({\em e.g.}, re-scaling, flipping). 
Among them, geometric operations, such as multi-scale training, improve scale robustness~\cite{fasterrcnn,detectron}.
Several hand-crafted data augmentation strategies were developed to improve performance and robustness of the detector~\cite{snip,sniper}.
Previous work~\cite{instaboost,dwibedi2017cut} also improves box-level augmentation by enriching foreground data.
Though inspiring performance gain has achieved,
these data augmentation strategies usually rely on heavy expert experience.

Automatic data augmentation policies were widely explored
in image classification~\cite{tran2017bayesian,zhu2017data,ratner2017learning,perez2017effectiveness,autoaug}. 
Its potential for object detection, however, was not thoroughly released.
One attempt to automatically learn data augmentation policies for object detectors is AutoAug-det~\cite{autoaug-det}\footnote{We refer it as AutoAug-det~\cite{autoaug-det} to distinguish from AutoAugment~\cite{autoaug}.}, which performs color or geometric augmentation upon the context of boxes. It does not fully consider the scale issue from {\em image-} and {\em box-level}, which are found, however, essential in object detector design~\cite{snip,sniper,instaboost}.
Moreover, the heavy computational search cost ({\em i.e.}, 400 TPU for 2 days) impedes it from vastly practical.
Thus, scale-aware property and efficiency issue are essential to address for searching augmentation in box-level tasks.

In this paper, we propose a new way to automatically learn scale-aware data augmentation strategies for object detection and
relevant
box-level tasks.
We first introduce scale-awareness to the search space from two {image-} and {box-levels}.
For image-level augmentations, zoom-in and -out operations are included with their probabilities and zooming ratios for search. For box-level augmentations, the augmenting areas are generalized with a new searchable parameter, i.e., area ratio. This makes box-level augmentations adaptive to object scales.

Based on our scale-aware search space, we further propose a new estimation metric to facilitate the search process with better efficiency. Previously, each candidate policy is estimated by the validation accuracy on a proxy task~\cite{autoaug,fastautoaug}, which lacks efficiency and accuracy to an extend. Our metric takes advantage of more specific statistics, that is, validation accuracy and accumulated loss over different scales, {\em to measure the scale balance}.
We empirically show that it yields a clearly higher correlation coefficient with the actual accuracy than the previous proxy accuracy metric.

The proposed approach is distinguished from previous work from two aspects.
First, different from hand-crafted policies, the proposed method utilizes automatic algorithms to search among a large variety of augmentation candidates.
It is hard to be fully explored or achieved by human effort.
Moreover, compared with previous learning-based methods, 
our approach fully explores the important scale issue in both image-level and box-level. With the proposed search space and evaluation metric, our method attains decent performance with much ({\em i.e.}, 40$\times$) less search cost.

The overall approach, called {\em \Ours}, can be easily instantiated for box-level tasks, which will be elaborated on in Sec.~\ref{sec:main_autoaug}. To validate its effectiveness, we conduct extensive experiments on MS COCO and Pascal VOC dataset~\cite{coco,PascalVOC} with several anchor-based and anchor-free object detectors, which are reported in Sec.~\ref{sec:abla_study}.

In particular, with ResNet-50 backbone, the searched augmentation policies contribute non-trivial gains over the strong MS baseline of RetinaNet~\cite{retinanet}, Faster R-CNN~\cite{fasterrcnn}, and FCOS~\cite{tian2019fcos}, and achieve 41.3\% AP, 41.8\% AP, and 42.6\% AP, respectively. We further experiment with more box-level tasks, like instance segmentation and keypoint detection. Without bells-and-whistles, our improved FCOS model attains 51.4\% AP with the search augmentation policies. Besides, our searched policies present meaningful patterns, which provide intuitive insight for human knowledge.

\begin{figure*}[ht]
\begin{center}
   \includegraphics[width=\linewidth]{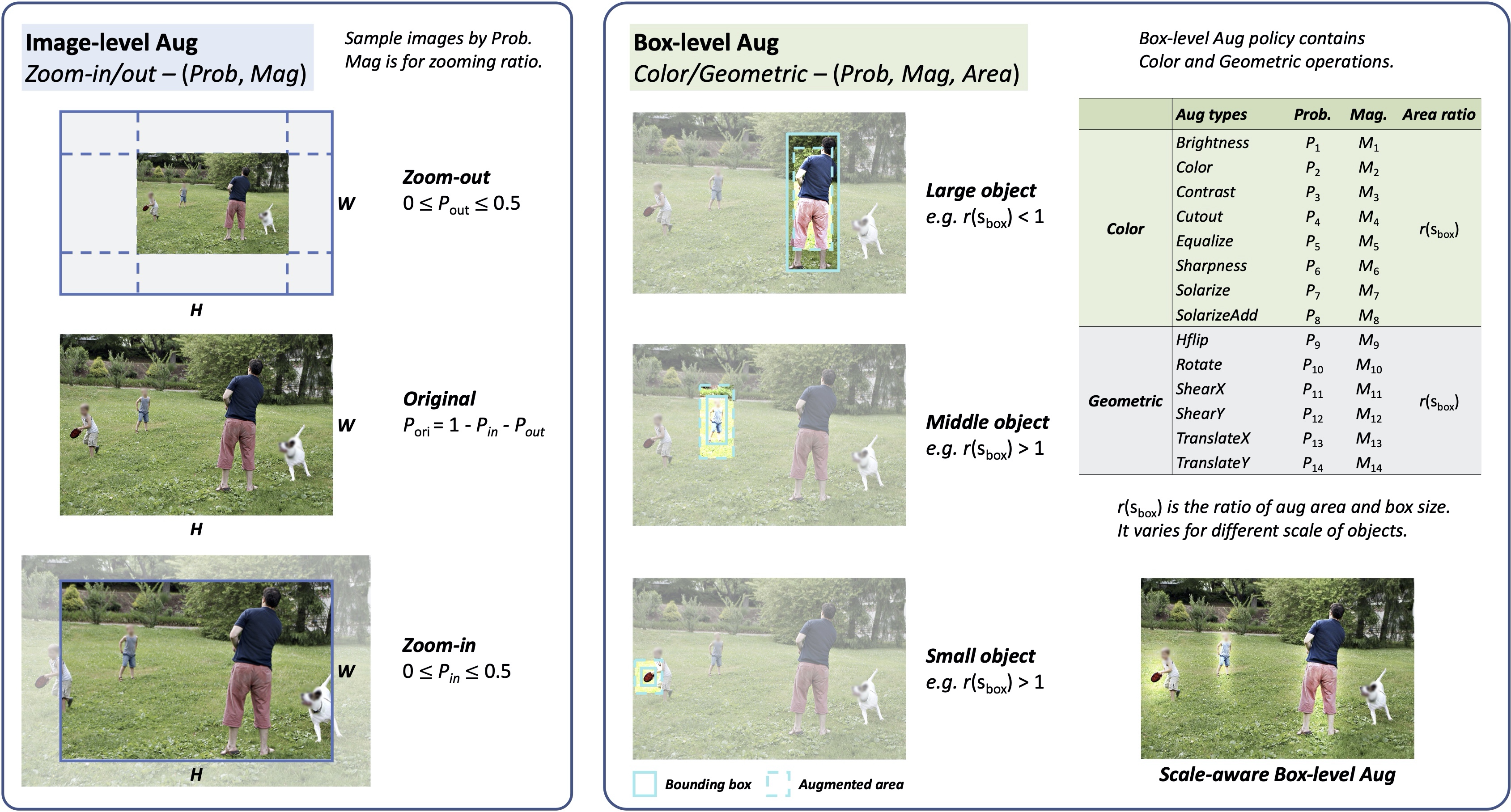}
   \caption{\small Scale-aware search space. It contains image-level and box-level augmentation. Image-level augmentation includes zoom-in and zoom-out functions with probabilities and magnitudes for search. In box-level, we introduce scale-aware area ratios, which make operations adaptive to objects in different scales. Augmented images are further generalized with the Gaussian map.}
   \label{fig:searchspace}
\end{center}
\end{figure*}

\section{Related Work}

Data augmentation has been widely utilized for network optimization and proven to be beneficial in vision tasks~\cite{imagenet, vgg, fasterrcnn, ssd, fcn, amodal}. Traditional approaches could be roughly divided into color operations ({\em e.g.,} brightness, contrast, and whitening) and geometric operations ({\em e.g.,} scaling, flipping, translation, and shearing), which require hyper-parameter tuning and are usually task-specific~\cite{liu2020deep}. Some commonly used strategies on image classification include random cropping, image mirroring, color shifting/whitening~\cite{krizhevsky2012imagenet}, Cutout~\cite{cutout}, and Mixup~\cite{mix-up}.

Scale-wise augmentations also play a vital role for the optimization of object detectors~\cite{wu2019detectron2,mmdetection}.
For example, SNIPER~\cite{sniper} generates image crops around ground truth instances with multi-scale training. 
YOLO-v4~\cite{yolov4} and Stitcher~\cite{stitcher} introduce mosaic inputs that contain re-scaled sub-images. 
For box-level augmentation, Dwibedi et al.~\cite{dwibedi2017cut} improve detection performance with the cut-and-paste strategy. And the visual context surrounding objects are modeled in~\cite{dvornik2018modeling}. Furthermore, InstaBoost~\cite{instaboost} augments training images using annotated instance masks with a location probability map. However, these hand-crafted designs still highly rely on expert efforts. 

Inspired by recent advancements in neural architecture search (NAS)~\cite{zoph2016neural, zoph2017learning, Real2018Regularized, detnas}, researchers try to learn augmentation policies from data automatically. An example is AutoAugment~\cite{autoaug}, which searches data augmentations for image classification and achieves promising results. 
PBA~\cite{pba} uses population-based search method for better efficiency. 
Fast AutoAugment~\cite{fastautoaug} applies Bayesian optimization to learn data augmentation policies. RandAug~\cite{randaug} removes the search process at the price of manually tailoring the search space to a very limited volume.
AutoAug-Det~\cite{autoaug-det} extends AutoAugment~\cite{autoaug} to object detection by taking box-level augmentations into consideration. 

\section{Scale-aware AutoAug}~\label{sec:main_autoaug}
In this section, we first briefly review the auto augmentation pipeline. 
Then, the {\em scale-aware search space} and {\em estimation metric} will be respectively elaborated in Sec.~\ref{sec:search_space} and Sec.~\ref{sec:search_metric}. 
We 
finally show
the search framework in Sec.~\ref{sec:search_framework}.

\subsection{Review of AutoAug}~\label{sec:brief_review}
Auto augmentation methods~\cite{autoaug, autoaug-det, pba, fastautoaug, differentiableautoaug} commonly formulate the process of finding the best augmentation policy as a search problem. 
To this end, three main components are needed, namely the {\em search space}, {\em search algorithm}, and {\em estimation metric}. 
Search space may vary according to tasks. For example, the search space~\cite{autoaug, pba, fastautoaug} is developed to image classification, while it is not the specified case for box-level tasks. 
As for search algorithms, reinforcement learning~\cite{zoph2016neural} and evolutionary algorithm~\cite{Real2018Regularized} are usually utilized to explore the search space in iterations.
During this procedure, each child model, which is optimized with the searched policy $p$, is evaluated on a designed metric to estimate its effectiveness. 
This metric serves as feedback for the search algorithm. 

\subsection{Scale-aware Search Space}~\label{sec:search_space}
The designed scale-aware search space contains both {\em image-level} and {\em box-level} augmentations. The image-level augmentations include zoom-in and zoom-out functions on the whole image. 
As for box-level augmentations, color and geometric operations are searched for objects in images.

\vspace{0.5em}
\noindent
\textbf{Image-level augmentations.}
To handle scale variations, object detectors are commonly trained with image pyramids. However, these scale settings highly rely on hand-crafted selection. In our search space, we alleviate this burden by searchable zoom-in and zoom-out functions. As illustrated in the left part of Fig.~\ref{fig:searchspace}, zoom-in and zoom-out functions are specified by probabilities $P$ and magnitudes $M$.
Specifically, the probabilities $P_\mathrm{in}$ and $P_\mathrm{out}$ are searched in the range from 0 to 0.5. With this range, the existence of original scale could be guaranteed with the probability
\begin{equation}
    P_\mathrm{ori} = 1 - P_\mathrm{out} - P_\mathrm{in}.
\end{equation}

The magnitude $M$ represents the zooming ratio for each function. 
For the zoom-in function, we search a zooming ratio from 0.5 to 1.0. For the zoom-out function, we search a zooming ratio from 1.0 to 1.5. For example, if a zooming ratio of 1.5 is selected, it means that the input images might be increased by $1.5\times$. In traditional multi-scale training, large-scale images would introduce an additional computational burden. To avoid this issue, we reserve the original shape in the zoom-in function with random cropping. 

After the search procedure, input images are randomly sampled from zoom-in, zoom-out, and original scale images with the searched $P$ and $M$ in each training iteration. In other words, we sample from 3 resolutions, a larger one, a small one and the original with the searched probabilities, {\em i.e.}, \{$P_{in}, P_{out}, P_{ori}$\}.
To our best knowledge, no previous work considers automatic scale-aware transformation search for object detection. Experiments validate the superiority over traditional multi-scale training in Tab.~\ref{tab:analysis-retinanet}. 

\vspace{0.5em}
\noindent
\textbf{Box-level augmentations.}
\begin{figure}[ht]
\vspace{-1.5em}
\begin{center}
    \subfigure[Comparison between square and gaussian transform.]{
   \includegraphics[width=0.9\linewidth]{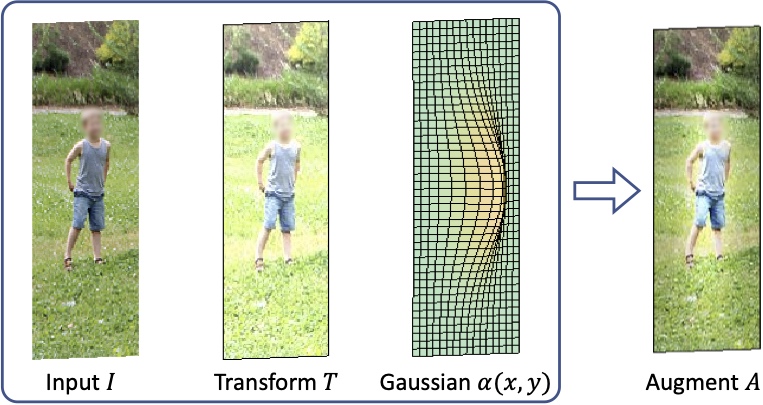}
   \label{fig:box_aug_0}
   }
   \vspace{-0.5em}
   \subfigure[Gaussian-based transform process.]{
   \includegraphics[width=0.9\linewidth]{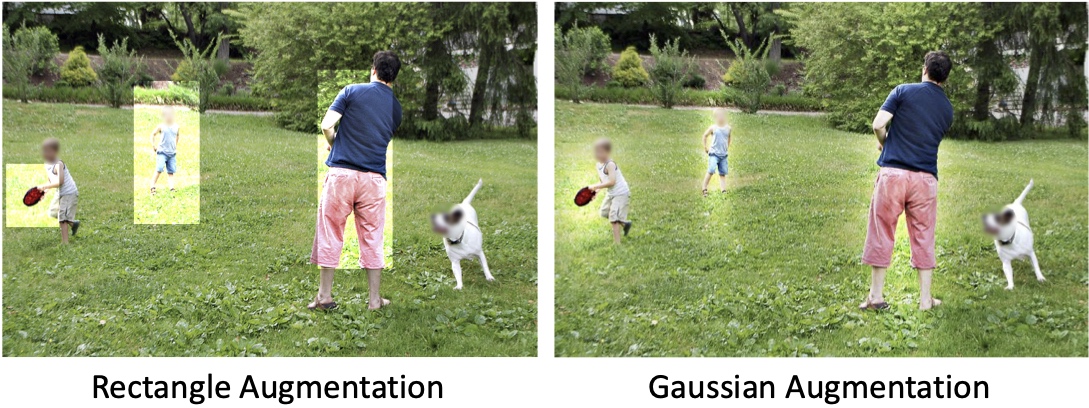}
   \label{fig:box_aug_1}
    }
\end{center}
\vspace{-0.5em}
\caption{\small An example of Gaussian-based box-level augmentation. It removes the original hard boundary and the augmented areas are adjustable to the Gaussian variance.}
\label{fig:box-level-augs}
\end{figure}
The box-level augmentations are designed to conduct transformation for each object box. Different from~\cite{autoaug-det}, the proposed approach further smooths the augmentations and relaxes it to contain learnable factors, {\em i.e.,} area ratio. 
In particular, previous box-level augmentation~\cite{autoaug-det} works exactly in the whole bounding box annotations without attenuation, which generate an obvious boundary gap between the augmented and original region. The sudden appearance change could reduce the difficulty for networks to locate the augmented objects, which brings the gap between training and inference.
To solve this issue, we extend the original rectangle augmentation to a gaussian-based manner. 
A visualization example of a box-level {\em color brightness} operation is given in Fig.~\ref{fig:box_aug_0}. 
We blend the original and the transformed pixels with a spatial-wise Gaussian map $\alpha(x,y)$ by
\begin{equation}
   A = \alpha(x,y) \cdot I + (1-\alpha(x,y)) \cdot T, \label{eq:blending}
\end{equation}
where $I$, $T$, and $A$ denotes the input, transformation function, and augmented region, respectively. This process is depicted in Fig.~\ref{fig:box_aug_1}. Actually, the designed gaussian-based process softens the boundary gap in a more natural manner. 

The second issue in previous operations is the lack of considering receptive fields and object scales. 
A common belief is that neural networks largely rely on the context information to recognize objects~\cite{upperboundfordet,receptivefields}.
Experimentally, we find that it may not be correct for objects in all scales, while the effect varies with objects scales. This is demonstrated with widely applied two-stage and one-stage detectors, namely the Faster R-CNN~\cite{fasterrcnn} and RetinaNet~\cite{retinanet}. As presented in Tab.~\ref{tab:context-scales}, if we test it on the COCO validation set with all context (background) pixels removed, its accuracy on small objects, AP$_s$, dramatically declines from 25.2\% to 18.0\%.
In contrast, AP$_l$ increases from 53.0\% to 56.1\%. It is consistent with that in RetinaNet~\cite{retinanet}.
This motivates us that augmentations merely inside/outside object boxes may not deal with objects in all scales appropriately. To this end, we introduce a searchable parameter, {\em area ratio}, which makes the aug area adaptive to object sizes. 
\begin{table}[t]
\begin{center}
\caption{\small Analysis on the context for scales. On well-trained ResNet-101 detectors, AP$_s$ drops and AP$_l$ increases consistently if contexts are removed in validation images.}
\resizebox{0.96\linewidth}{!}{
\begin{tabular}{|l|ccccc|}
\hline
                              & \multicolumn{1}{c|}{\em with context} & AP  & AP$_s$  & AP$_m$  & AP$_l$ \\ \hline
\multirow{3}{*}{Faster R-CNN} & \multicolumn{1}{c|}{\cmark}     & 41.4   & 25.2 &  44.8   & 53.0 \\
                              & \multicolumn{1}{c|}{\xmark}      & 40.5  & 18.0 & 45.7 &  56.1   \\ \cline{2-6}
                              & \multicolumn{1}{c}{$\Delta$} & \textcolor{red}{-0.9} & \textcolor{red}{\bf -7.2} & \textcolor[RGB]{34,139,34}{+0.9} & \textcolor[RGB]{34,139,34}{\bf +3.1}\\ \hline
\multirow{3}{*}{RetinaNet}    & \multicolumn{1}{c|}{\cmark}     &  40.3   & 23.3 & 44.0 &  53.3   \\
                              & \multicolumn{1}{c|}{\xmark}      & 39.8   &  16.7   & 44.4 & 57.7 \\ \cline{2-6}
                              & \multicolumn{1}{c}{$\Delta$}  & \textcolor{red}{-0.5}   & \textcolor{red}{\bf -6.6} & \textcolor[RGB]{34,139,34}{+0.4} & \textcolor[RGB]{34,139,34}{\bf +4.4} \\ \hline
\end{tabular}
}
\vspace{-1.5em}
\label{tab:context-scales}
\end{center}
\end{table}

Here, we generalize the Gaussian map with the parameter of area ratio. Given an image with size $H\times W$ and bounding box annotations, the box ($x_c$, $y_c$, $h$, $w$) could be represented with the central point ($x_c$, $y_c$) and the height/width $h$/$w$. We formulate the Gaussian map by
\begin{equation}
    \alpha(x,y) = \exp\left(-\left(\frac{(x-x_c)^2}{2\sigma_x^2} + \frac{(y-y_c)^2}{2\sigma_y^2}\right)\right).
\end{equation}

Then, we define the augmentation area $V$ as the integration of the Gaussian map, where
\begin{equation}
    V = \int_{0}^{H} \int_{0}^{W} \alpha(x,y)\, \mathrm{d}x\mathrm{d}y.
\end{equation}

The {\em area ratio} for the box-level augmentation is denoted as $r$. 
Here, $r(s_{\mathrm{box}}) = V/s_{\mathrm{box}}$ is searchable for different scales, which determines the spatially augmentation area for each object.
Thus, the standard deviation factors, $\sigma_x$ and $\sigma_y$, could be calculated as in Eq.~\eqref{eq:sigma}. We provide the detailed calculation process in the {\em supplementary materials}.
\begin{equation}
    \sigma_x = h\sqrt{\frac{W/H}{2\pi}r}, \quad
    \sigma_y = w\sqrt{\frac{H/W}{2\pi}r}.
    \label{eq:sigma}
\end{equation}

\vspace{0.5em}
\noindent
\textbf{Search space summary.}
Our search space contains both image-level and box-level augmentations. 
For the image-level augmentation, we search for the parameters of zoom-in and zoom-out operations.
To keep consistent with the convention~\cite{autoaug-det}, our box-level operations have 5 sub-policies, where sub-policy consists of a {\em color} operation as well as a {\em geometric} operation. 
Each operation contains two parameters, namely the {\em probability} and {\em magnitude}. The probability is sampled from a set of 6 discrete values, from 0 to 1.0 with 0.2 as the interval. The magnitude represents the strength factor for each operation with custom range values.
We map the magnitude range to a standardized set of 6 discrete values, from 0 to 10 with 2 as the interval. For box-level operations, there are 3 area ratios to search for small, middle and large scales. Each area ratio is independently searched in a discrete set of 10 values. We list the details of these operations in the {\em supplementary materials}. In summary, the total search space provides {\small $(6^2)^2\times(((6\times6^2)\times(8\times 6^2))^5\times10^3)=1.2^{30}$} candidate policies, which is twice large as~\cite{autoaug-det}.

\subsection{Scale-aware Estimation Metric}~\label{sec:search_metric}
Autoaugment methods commonly employ validation accuracies on a proxy task (a small subset with training images) as the search metric. 
However, such a manner is found to be inaccurate and computationally expensive~\cite{randaug}, which will be further demonstrated in Fig.~\ref{fig:coefficient}. 
In contrast, all operations in our search space, both image-level and box-level, have explicit relationships with each scale. 
Thanks for the convenience, a scale-aware metric can be further proposed to capture more specific statistics of different scales. 
Specifically, the evaluation metric is established based on an observation that {\em balanced optimization over different scales would be beneficial to training}. Thus, a scale-aware metric can be formulated with the record of the accumulated loss and accuracy on different scales during fine-tuning. 

Given a plain model trained without data augmentation, we record its validation accuracy AP and accuracies on each scale $AP_i$ with $i\in S$.
For each candidate policy $p$, a child model is further fine-tuned upon it. 
We record the accumulated loss $L^p_i$, the validation accuracies on each scale $AP^p_i$, and the overall $AP$ to formulate the objective function as
\begin{equation}
    \min\limits_{p} f(\{L^p_{i\in S}\}, \{AP^p_{i\in S}\}).
    \label{eq:objectivefunction}
\end{equation}

Balanced optimization over different scales is essential to the performance and robustness of object detectors.
An intuitive way to measure the balance is the standard deviation $\sigma(\{L^p_i|i\in S\})$ of losses on various scales. However, we find it sometimes delves into a sub-optimal where other scales are sacrificed to achieve the optimization balance.

Here, we adopt the 
principle
of Pareto Optimality~\cite{black2012dictionary} to overcome this obstacle.
In particular, we introduce a concept, named Pareto Scale Balance, to describe our objective: {\em the optimization over scales can not be better without hurting the accuracy of any other scale}. 
To this end, we introduce a penalization factor $\Phi$ to punish the scales $\widehat{S}$ where accuracy drops after fine-tuning with the policy $p$. 
Therefore, the metric function can be upgraded to
\begin{equation}
    f(\{L^p_{i\in S}\}, \{AP^p_{i\in S}\}) = \sigma(\{L^p_{i\in S}\}) \cdot \Phi(\{AP^p_{i\in \widehat{S}}\}), 
    \label{eq:metric}
\end{equation}
where $\Phi(\{AP^p_{i\in \widehat{S}}\}) = \prod_{i\in \widehat{S}}\frac{AP_i}{AP^p_i}$ and $\frac{AP_i}{AP^p_i}$ is the scale-wise ratio of the original and the fine-tuned accuracy.

Compared with previous proxy-accuracy metrics, ours is superior in {\em computational efficiency} and {\em estimation accuracy}. Towards efficient computation, child models are fine-tuned upon the given plain model, instead of training from scratch. We record the changes that resulted from the augmented fine-tuning to compute our metric.
For accurate estimation, more specific statistics, that is, AP and loss over various scales, is reasonable to receive a higher coefficient with the actual performance.
Experimentally, we carefully compare the proposed search metric with the original accuracy metric to verify the effectiveness in Sec.~\ref{sec:exp_search_metric}.

\begin{algorithm}[t]
    \caption{Search Framework} 
    \label{algo:search}
    \SetKwInOut{Input}{Input}\SetKwInOut{Output}{Output}
    \Input{Plain model $m$,
           Initialized Population $\mathbf{P}$,
           Training set $\mathbb{T}$,
           Val set $\mathbb{V}$, 
           Iterations \#$\mathrm{I}$.}
    \begin{small}
        \texttt{$f^*, p^*$ $\gets$ ($\infty, \varnothing$)}

        \For {$k$ $\in$ $(1\;to\;$\#$\mathrm{I}$)} {
            \For {$p$ $\in$ $\mathbf{P}$} {
                \texttt{$m_p$, $\{L^p_{i\in S}\}$ $\gets$ finetune($m$,$\mathbb{T}$,$\theta_{p}$)}
                
                \texttt{$\{AP^p_{i\in S}\}$ $\gets$ evaluate($m_p$, $\mathbb{V}$)}
                
                \texttt{$f_p$ $\gets$ $f(\{L^p_{i\in S}\}, \{AP^p_{i\in S}\})$ -- Eq.(\ref{eq:metric})}
                
                \If {$f_p$ $<$ $f^*$} {
                    \texttt{$f^*, p^*$ $\gets$ ($f_p, p$)}}
            }
            
            \texttt{$\mathcal{P}$ $\gets$ select-topk($\mathbf{P}$)}
            
            \texttt{$\mathbf{P}$ $\gets$ mutate-crossover($\mathcal{P}$)}
        }
    \end{small}
    
    \Output{The best augmentation policy $p^*$}
\end{algorithm}


\begin{table}[t]
\begin{center}
\caption{\small Improvement details on RetinaNet ResNet-50.}
\vspace{-6pt}
\resizebox{\linewidth}{!}{
\begin{tabular}{|l|c|ccccc|}
\hline
                & AP                       & AP$_{50}$                     & AP$_{75}$                     & AP$_s$                      & AP$_m$                      & AP$_l$                      \\
                                   \hline
MS Baseline              & 38.2 & 57.3 & 40.5 & 23.0 & 41.6 & 50.3  \\
Ours image-level    &  40.1  & 59.8  & 43.3  & 24.0  & 44.1  &  53.1  \\
+ box-level  & 40.6 & 60.4 & 43.6 & 24.1	& 44.4 & 53.5 \\
+ scale-aware area  & \textbf{41.3} & \textbf{61.0} & \textbf{44.1} & \textbf{25.2} & \textbf{44.5} & \textbf{54.6} \\ \hline
\end{tabular}}
\vspace{-1.5em}
\label{tab:analysis-retinanet}
\end{center}
\end{table}
\begin{table}[t]
\begin{center}
\caption{\small Comparison with AutoAug-det on RetinaNet ResNet-50.}
\vspace{-0.5em}
\resizebox{\linewidth}{!}{
\begin{tabular}{|l|c|c|ccc|}
\hline
   & search cost & AP &  AP$_s$ &  AP$_m$ & AP$_l$              \\
                                   \hline
AutoAug-det~\cite{autoaug-det} & 800 TPU-days & 36.7$\to$39.0 &- & -&- \\ 
AutoAug-det$^\dag$~\cite{autoaug-det}\tablefootnote{$^\dag$ means our implementation with the same baseline settings to ours.} & 800 TPU-days  & 38.2$\to$40.3 & 23.6 & 43.9 & 53.8 \\ 
Ours  & 20 GPU-days & 38.2$\to$\textbf{41.3}  & 25.2 & 44.5 & 54.6 \\ 
 \hline
\end{tabular}}
\vspace{-1.5em}
\label{tab:compare-autoaugdet}
\end{center}
\end{table}
\begin{table}[t]
\begin{center}
\caption{\small Search on RetinaNet ResNet-50 with different metrics.}
\vspace{-0.5em}
\resizebox{\linewidth}{!}{
\begin{tabular}{|l|c|ccccc|}
\hline
                           & AP   & AP$_{50}$ & AP$_{75}$ 
                           & AP$_s$  & AP$_m$  & AP$_l$  
                           \\ \hline
proxy accuracy in~\cite{autoaug-det}& 40.0 & 59.7 & 42.5 & 23.9 & 44.1 & 52.6 
\\
scale loss std $\sigma$  & 40.7 & 60.5 & 43.5 & 24.1 & 44.5 & 53.5 
\\
our metric - Eq.~(\ref{eq:metric})                       & \textbf{41.3} & \textbf{61.0} & \textbf{44.1} & \textbf{25.2} & 44.5 & \textbf{54.6} 
\\ \hline
\end{tabular}
}
\vspace{-1.5em}
\label{tab:ablation-search-metrics}
\end{center}
\end{table}

\subsection{Search Framework}~\label{sec:search_framework}
Given the above search space and search metric, we describe the search framework in this section. 
In this work, the evolutionary algorithm, {\em e.g.,} tournament selection algorithm~\cite{Real2018Regularized}, is adopted as the search controller. 
Specifically, a population of $|P|$ policies are sampled from the search space in each iteration. 
After evaluating the sampled policies, we select the top $k$ policies as {\em parent} for the next generation. Then, child policies are produced by mutation and crossover among the parent policies. This process is repeated for iterations until convergence.

To evaluate augmentation policies, we 
first
train a plain model with no data augmentation. Then, we fine-tune it upon each augmentation policy for $n$ iterations and record the accumulated loss during optimization. We also record its accuracy {\em before} and {\em after} fine-tuning. With these statistics, the search metric for each policy could be obtained.
This search framework is illustrated in Alg.~\ref{algo:search}.
\begin{table}[t]
\begin{center}
\caption{\small Improvements across detection frameworks.}
\vspace{-0.5em}
\resizebox{\linewidth}{!}{
\begin{tabular}{|l|l|c|ccccc|}
\hline
Models & {\em policy}     & AP                       & AP$_{50}$                     & AP$_{75}$                     & AP$_s$                      & AP$_m$                      & AP$_l$  \\ \hline \hline
\multicolumn{8}{|l|}{\em RetinaNet:}\\ \hline
\multirow{3}{*}{\small ResNet-50\tablefootnote{FPN~\cite{fpn} is used as a default setting, unless -C4 is denoted.}}  & {\small Baseline}    & 36.6 & 55.7 & 39.1 & 20.8 & 40.2 & 49.4 \\
                            & {\small MS Baseline} & 38.2 & 57.3 & 40.5 & 23.0 & 41.6 & 50.3 \\
                             & Ours      & \textbf{41.3} & 61.0 & 44.1 & 25.2 & 44.5 & 54.6 \\ \hline
\multirow{3}{*}{\small ResNet-101} & {\small Baseline}    & 38.8 & 59.1 & 42.3 & 21.8 & 42.7 & 50.2    \\ 
                             & {\small MS Baseline} & 40.3 & 59.8 & 42.9 & 23.2 & 44.0 & 53.2 \\
                             & Ours      & \textbf{43.1} & 62.8 & 46.0 & 26.2 & 46.8 & 56.7 \\ \hline \hline
\multicolumn{8}{|l|}{\em Faster R-CNN:}\\ \hline
\multirow{3}{*}{\small ResNet-50}  & {\small Baseline}    & 37.6    & 57.8 & 41.0 & 22.2 & 39.9 & 48.4 \\
  & {\small MS Baseline} & 39.1 & 60.8 &  42.6 &  24.1 &  42.3 &  50.3 \\
                             & Ours      & \textbf{41.8}    & 63.3 & 45.7 & 26.2 & 44.7 & 54.1 \\ \hline 
\multirow{3}{*}{\small ResNet-101} & {\small Baseline}    & 39.8    &   61.3   & 43.5  & 23.1 & 43.2 & 52.3 \\
& {\small MS Baseline} & 41.4  & 60.4  & 44.8  & 25.0  & 45.5  & 53.1  \\
                             & Ours      & \textbf{44.2}    & 65.6 & 48.6 & 29.4 & 47.9 & 56.7 \\ \hline \hline
\multicolumn{8}{|l|}{\small \em FCOS:} \\ \hline
\multirow{2}{*}{\small ResNet-50}  
 & {\small MS Baseline} & 40.8 & 59.6 & 43.9 & 26.2 & 44.9 & 51.9 \\
                             & Ours      & \textbf{42.6} & 61.2 & 46.0 & 28.2 & 46.4 & 54.3 \\ \hline
\multirow{2}{*}{\small ResNet-101} 
 & {\small MS Baseline} & 41.8 & 60.3 & 45.3 & 25.6 & 47.7 & 56.1 \\ 
                             & Ours      & \textbf{44.0} & 62.7 & 47.3 & 28.2 & 47.8 & 56.1 \\ \hline
\end{tabular}}
\vspace{-1.5em}
\label{tab:improves-frameworks}
\end{center}
\end{table}

\begin{table}[t]
\begin{center}
\caption{\small Improvements across tasks on Mask R-CNN.}
\resizebox{\linewidth}{!}{
\begin{tabular}{|l|l|ccc|ccc|}
\hline
Models                & {\em policy}    & AP$^{\textrm{m/k}}$                       & AP$_{50}^{\textrm{m/k}}$                     & AP$_{75}^{\textrm{m/k}}$                     & AP$^{\textrm{b}}$                      & AP$_{50}^{\textrm{b}}$                      & AP$_{75}^{\textrm{b}}$  \\ \hline \hline
\multicolumn{8}{|l|}{\em Instance Segmentation:} \\ \hline
\multirow{2}{*}{ResNet-50}   & {\small MS Baseline} & 36.4 & 58.8 & 38.7 & 40.4 & 61.9 & 44.0 \\
                             & Ours      & \textbf{38.1} & 60.9 & 40.8 & \textbf{42.8} & 64.4 & 46.9 \\ \hline
\multirow{2}{*}{ResNet-101}  & {\small MS Baseline} & 37.9 & 60.4 & 40.4 & 42.3 & 63.8 & 46.6 \\
                             & Ours      & \textbf{40.0} & 63.2 & 42.9 & \textbf{45.3} & 66.4 & 49.8 \\ \hline \hline
\multicolumn{8}{|l|}{\em Keypoint Estimation:} \\ \hline
\multirow{2}{*}{ResNet-50}  & {\small MS Baseline}   & 64.1 & 85.9 & 69.7 & 53.5 & 82.7 & 58.4 \\
                             & Ours      & \textbf{65.7} & 86.6 & 71.7 & \textbf{55.5} & 84.2 & 60.9 \\ \hline
\multirow{2}{*}{ResNet-101} & {\small MS Baseline}    & 65.1 & 86.5 & 71.2  & 54.8  & 83.2 & 60.0 \\
                             & Ours      & \textbf{66.4} & 87.5 & 72.7 & \textbf{56.5} & 84.6 & 62.1\\ \hline
\end{tabular}}
\vspace{-1.5em}
\label{tab:improves-maskrcnn}
\end{center}
\end{table}
\section{Experiments}
\label{sec:experiments}

\subsection{Implementation Details}
\vspace{0.5em}
\noindent
\textbf{Policy search.}
In the search phase, we adopt RetinaNet~\cite{retinanet} on ResNet-50~\cite{he2016deep} backbone. We split the detection dataset into a training set for child model training, a validation set for evaluation during search, and the test set \texttt{val2017} for final evaluation. The validation set contains 5k images randomly sampled from the \texttt{train2017} in MS COCO~\cite{coco} and the remains are for child model training. Each child model is fine-tuned for 1k iterations on the plain model, which is just an arbitrary partially trained baseline model. In the evolutionary search, the evolution process is repeated for 10 iterations. The evolution population size is 50 and the top 10 models are selected as subsequent parents.

\vspace{0.5em}
\noindent
\textbf{Final policy evaluation.}
Models are trained with the searched augmentation policy in the typical pre-training and fine-tuning schedule on MS COCO dataset. The training images are resized such that the shorter size is 800 pixels. Faster R-CNN and RetinaNet models are trained for 540k iterations to fully show its potential, while others are trained for 270k iterations. 
Multi-scale training baselines are enhanced by randomly selecting a scale between 640 to 800 during training. 
We train models on 8 GPUs with a total 16 images per batch. The learning rate is initialized as 0.02. We set weight decay as 0.0001 and momentum as 0.9.

\subsection{Verification}~\label{sec:abla_study}
In this section, we systematically evaluate our proposed \Ours. We first present the improvements from the search policy on the target task and then show its transferability to other tasks and datasets. After that, we analyze the proposed search metric in detail.

\vspace{0.5em}
\noindent
\textbf{Improvements analysis.}
The top block in Tab.~\ref{tab:improves-frameworks} shows the improvements from our searched augmentation policy on RetinaNet. 
On ResNet-50 backbone, our searched augmentation policy enhances the competitive multi-scale training baseline to 41.3\% AP by 3.1\%. On ResNet-101, it achieves a 2.8\% gains to 43.1\% AP. 
We also perform experiments upon the large scale jittering~\cite{spinenet} in the {\em supplementary materials}. 
These improvements come from training data augmentations and introduce no additional cost to inference.

For a better understanding of the improvements, we show the component-wise improvements in Tab.~\ref{tab:analysis-retinanet}. The image-level augmentations boost the performance by 1.9\% AP from 38.2\% to 40.1\%. Upon this, the non-scale-aware box-level augmentations improve the performance to 40.6\%. If it is further upgraded to be scale-aware, the performance gets an additional 0.7\% enhancement to 41.3\%. In contrast, in AutoAug-det~\cite{autoaug-det}, the box-level augmentations yield only 0.4\% improvements. 
The improvements mostly come from small and large objects, 
which verifies the effectiveness of scale-aware box-level augmentations.

In addition, we compare with the previous state-of-the-art auto augmentation method in Tab.~\ref{tab:compare-autoaugdet}. 
On RetinaNet~\cite{retinanet} with ResNet-50~\cite{he2016deep} backbone, AutoAug-det~\cite{autoaug-det} improves the baseline from 36.7\% to 39.0\% by 2.3\% AP. 
For better comparison, we implement the searched policy in AutoAug-det~\cite{autoaug-det} on our baseline. It is trained on the exact same settings, except for the data augmentation policy. 
It improves the baseline from 38.2\% to 40.3\% by 2.1\% AP. It is inferior to our +3.1\% improvement.
For small objects, our searched policy gets a more balanced performance ({\em i.e.}, +1.6 AP$_s$) thanks to the scale-aware search space and metric.
In terms of search cost, the data augmentation policy in AutoAug-det~\cite{autoaug-det} costs 800 TPU-days (400 TPUs on 2 days) for search, while our search costs only 8 GPUs (Tesla-V100) on 2.5 days. It is a 40$\times$ computational saving, without considering the machine type difference.

\begin{table*}[t]
\begin{center}
\caption{\small Improvements on PASCAL VOC with Faster R-CNN on ResNet-50 backbone.}
\vspace{-0.5em}
\resizebox{\linewidth}{!}{
\begin{tabular}{|l|c|cccccccccccccccccccc|}
\hline
          & mAP & plane & bike & bird & boat & bottle & bus & car & cat & chair & cow & table & dog & horse & mbike & person & plant & sheep & sofa & train & tv \\ \hline
baseline      &  78.6   & 80.9 & 80.8 & 79.3 & 72.3 & 67.2 & 87.4 & 88.5 & 88.6 & 62.6 & 86.0 & 71.2 & 88.0 & 88.9 & 80.6 & 79.9 & 52.6 & 78.7 & 74.0 & 86.2 & 78.3 \\
+ ours  &  \textbf{81.6}   & 88.7 & 88.2 & 80.1 & 74.1 & 73.6 & 88.3 & 89.1 & 88.9 & 68.1 & 87.2 & 73.8 & 88.4 & 88.9 & 87.5 & 87.1 & 56.2 & 79.0 & 79.7 & 87.2 & 78.6 \\ \hline
\end{tabular}}
\vspace{-1.5em}
\label{tab:pascal-voc}
\end{center}
\end{table*}

\begin{table*}[t]
\begin{center}
\caption{\small Comparison with state-of-the-art data augmentation methods for object detection.}
\resizebox{\linewidth}{0.3\linewidth}{
\begin{tabular}{|l|c|c|lccccc|}
\hline
Method          & Detector     & Backbone        & AP   & AP$_{50}$ & AP$_{75}$ & AP$_s$  & AP$_m$  & AP$_l$  \\ \hline \hline
{\em Hand-crafted:}   &              &                 &         &      &      &      &      &      \\
~~~~Dropblock~\cite{dropblock}       & RetinaNet    & ResNet-50       & 38.4    & 56.4 & 41.2 & -    & -    & -    \\
~~~~Mix-up~\cite{mix-up}          & Faster R-CNN & ResNet-101      & 41.1    & -    & -    & -    & -    & -    \\
~~~~PSIS$^{*}$~\cite{psis}            & Faster R-CNN & ResNet-101      & 40.2    & 61.1 & 44.2 & 22.3 & 45.7 & 51.6 \\
~~~~Stitcher~\cite{stitcher}            & Faster R-CNN & ResNet-101  & 42.1  & - & - & 26.9 & 45.5 & 54.1 \\
~~~~GridMask~\cite{gridmask}        & Faster R-CNN & ResNeXt-101     & 42.6    & 65.0 & 46.5 & -    & -    & -    \\
~~~~InstaBoost$^{*}$~\cite{instaboost}      & Mask R-CNN   & ResNet-101      & 43.0    & 64.3 & 47.2 & 24.8 & 45.9 & 54.6 \\ 
~~~~SNIP (MS test)$^{*}$~\cite{snip}            & Faster R-CNN & ResNet-101-DCN-C4  & 44.4    & 66.2 & 49.9 & 27.3 & 47.4 & 56.9 \\
~~~~SNIPER (MS test)$^{*}$~\cite{sniper}          & Faster R-CNN & ResNet-101-DCN-C4  & 46.1   & 67.0 & 51.6 & 29.6 & 48.9 & 58.1 \\
\hline \hline
{\em Automatic:} &              &                 &         &      &      &      &      &      \\
~~~~AutoAug-det~\cite{autoaug-det} & RetinaNet    & ResNet-50       & 39.0    &   -   &    -  & -  &   -   &  - \\
~~~~AutoAug-det~\cite{autoaug-det} & RetinaNet    & ResNet-101      & 40.4    &   -   &    -  & -  &   -   &  - \\
~~~~AutoAug-det$^\dag$~\cite{autoaug-det} & RetinaNet    & ResNet-50      & 40.3    &  60.0  &  43.0    &  23.6    &  43.9    &  53.8    \\ 
~~~~AutoAug-det$^\dag$~\cite{autoaug-det} & RetinaNet    & ResNet-101      & 41.8    &  61.5  &  44.8    &  24.4    &  45.9    &  55.9   \\ 
~~~~RandAug~\cite{randaug} & RetinaNet    & ResNet-101      & 40.1    &   -   &    -  & -  &   -   &  - \\
~~~~RandAug$^\dag$~\cite{randaug} & RetinaNet    & ResNet-101      & 41.4    &   61.4   &   44.5  & 25.0  &   45.4   &  54.2 \\
\hline \hline
{\em Ours:} &              &                 &         &      &      &      &      &      \\
~~~~Scale-aware AutoAug            & RetinaNet    & ResNet-50       & 41.3    & 61.0 & 44.1 & 25.2 & 44.5 & 54.6 \\
~~~~Scale-aware AutoAug            & RetinaNet    & ResNet-101      & 43.1    & 62.8 & 46.0 & 26.2 & 46.8 & 56.7 \\
~~~~Scale-aware AutoAug            & Faster R-CNN & ResNet-101      & 44.2    & 65.6 & 48.6 & 29.4 & 47.9 & 56.7 \\
~~~~Scale-aware AutoAug (MS test)  & Faster R-CNN & ResNet-101-DCN-C4  & 47.0 & 68.6 & 52.1 & 32.3 & 49.3 & 60.4 \\
~~~~Scale-aware AutoAug            & FCOS$\,\,$        & ResNet-101      & 44.0    & 62.7 & 47.3 & 28.2 & 47.8 & 56.1 \\
~~~~Scale-aware AutoAug            & FCOS$^{\ddag}$       & ResNeXt-32x8d-101-DCN & 48.5 &  67.2 & 52.8 &  31.5  & 51.9  &  63.0 \\ 
~~~~Scale-aware AutoAug (1200 size) & FCOS$^{\ddag}$         & ResNeXt-32x8d-101-DCN & 49.6 &  68.5 & 54.1 &  35.7  & 52.5  &  62.4 \\ 
~~~~Scale-aware AutoAug  (MS test) & FCOS$^{\ddag}$         & ResNeXt-32x8d-101-DCN & \textbf{51.4} &   69.6   &   57.0   &   37.4   &   54.2   &   65.1   \\ \hline
\end{tabular}}
\vspace{-1.5em}
\label{tab:comparison}
\end{center}
\end{table*}

\vspace{0.5em}
\noindent
\textbf{Transferability.}
Although our data augmentation policy is searched in object detection on RetinaNet, we make comprehensive experiments to show its effectiveness to work on other object detectors, datasets and relevant tasks.

In object detection, we verify our policy on mainframe anchor-based one-stage, two-stage, and anchor-free detectors. In addition to the previous RetinaNet experiments, we show our results on Faster R-CNN and FCOS in Tab.~\ref{tab:improves-frameworks}. The improvements on Faster R-CNN are remarkable, {\em i.e.}, +2.7\% and +2.8\% on ResNet-50 and ResNet-101, respectively. On the anchor-free detector FCOS, it achieves 44.0\% AP on ResNet-101 with similar improvements.\blfootnote{$^{\ddag}$ For FCOS ResNeXt-32x8d-101-DCN models, it is an improved version with ATSS~\cite{atss} for performance boosting. $^{*}$ results on \texttt{test-dev}. Our comparisons on \texttt{test-dev} are in the {\em supplementary materials}.}

Our augmentation policy is feasible in any box-level tasks. We validate its performance using Mask R-CNN~\cite{maskrcnn} on instance segmentation and keypoint estimation. Similar improvements are consistently present in Tab.~\ref{tab:improves-maskrcnn}. For instance segmentation, our Mask R-CNN model achieves 40.0\% mask AP on ResNet-101 backbone. 
In addition, we also transfer our augmentation policy to PASCAL VOC dataset. We train a Faster R-CNN model on ResNet-50 for 48k iterations and divide the learning rate at 36k iterations. It improves the baseline by 3\% mAP as in Tab.~\ref{tab:pascal-voc}.

\begin{table}[ht]
\begin{center}
\vspace{-0.5em}
\caption{\small Searched augmentation policy.}
\vspace{-1em}
\resizebox{\linewidth}{!}{
\begin{tabular}{|l|ll|}
\hline
{\em Image-level}   & (Zoom-in, 0.2, 4)      & (Zoom-out, 0.4, 10)     \\ \hline
{\em Box-level}     & {\em Color operations}  & {\em Geometric  operations}  \\
Sub-policy 1. & (Color, 0.4, 2)         & (TranslateX, 0.4, 4)   \\
Sub-policy 2. & (Brightness, 0.2, 4)    & (Rotate, 0.4, 2)       \\
Sub-policy 3. & (Sharpness, 0.4, 2)     & (ShearX, 0.2, 6)       \\
Sub-policy 4. & (SolarizeAdd, 0.2, 2)   & (Hflip, 0.3, 0)               \\
Sub-policy 5. & Original                & (TranslateY, 0.2, 8)   \\
Area ratio   & \multicolumn{2}{c|}{Small - 6 $\;\;$ Middle - 2 $\;\;$ Large - 0.4} \\ \hline
\end{tabular}}
\vspace{-3em}
\label{tab:Searched-policy}
\end{center}
\end{table}

\begin{figure}[t]
\begin{center}
   \includegraphics[width=\linewidth]{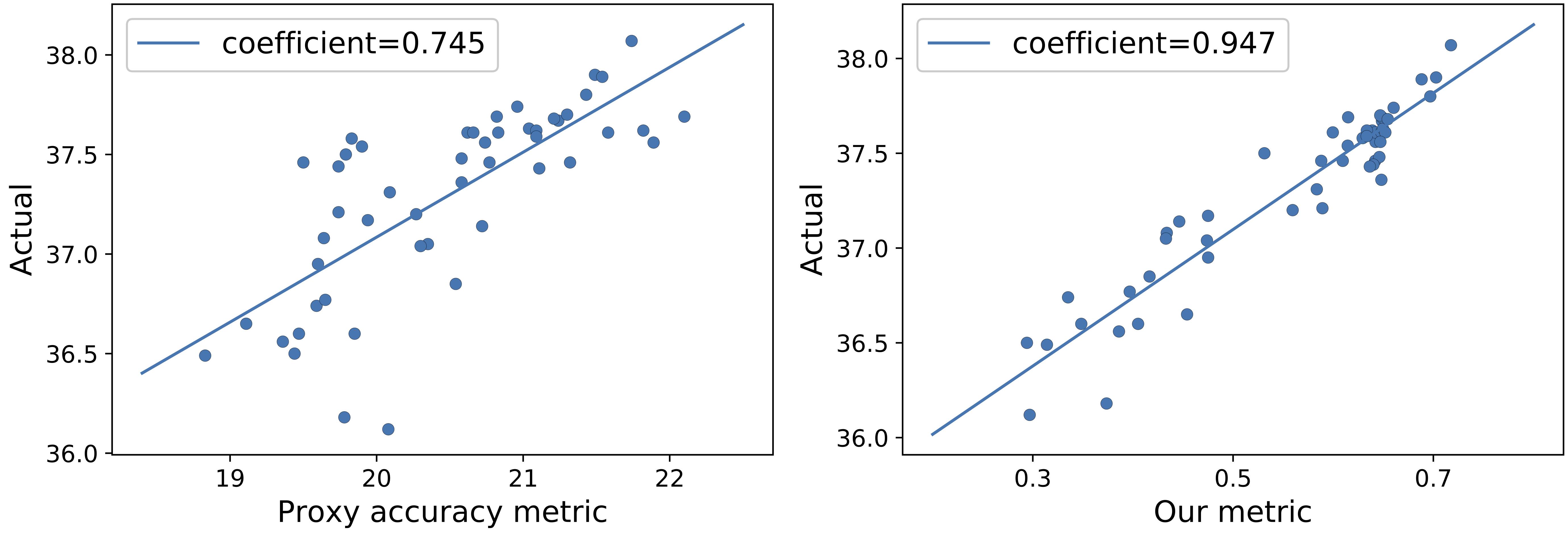}
   \caption{\small Coefficients between actual accuracy and metrics. Our metric presents a higher coefficient than the proxy accuracy~\cite{autoaug-det}.}
   \label{fig:coefficient}
\end{center}
\vspace{-1.5em}
\end{figure}

\vspace{0.5em}
\noindent
\textbf{Search metric analysis.}
\label{sec:exp_search_metric}
We compare our search metric with the proxy accuracy metric. For the proxy accuracy metric in~\cite{autoaug-det}, each model is trained on a subset training set, 5k images. For each search metric, we train 50 models with policies randomly sampled in the search space. Each model is trained for 90k iterations and evaluated on \texttt{val2017} to obtain the actual accuracy. Meanwhile, the proxy accuracy metric and our std-based metric are computed for each model. We illustrate the Pearson coefficients in Fig.~\ref{fig:coefficient}. Our std-based metric is horizontally flipped in [0, 1] for better illustration. It shows that our metric has a clearly higher coefficient to actual accuracies than the proxy accuracy metric. In addition, we use different metrics for search in Tab.~\ref{tab:ablation-search-metrics}. The search metric in~Eq.\eqref{eq:metric} is slightly better than the purely std metric, thanks to the penalty factor.

\subsection{Comparison}
We compare our final models on the MS COCO dataset with other data augmentation methods in object detection. The training settings are consistent with the implementation details mentioned before. As shown in Tab.~\ref{tab:comparison}, our augmentation method on Faster R-CNN with ResNet-101 backbone achieves 44.2\% AP, without any testing techniques. It is better than the augmentation methods with the same backbone, including InstaBoost~\cite{instaboost} on Mask R-CNN (43.0\% AP). To compare with the state-of-the-art hand-crafted augmentations, SNIP~\cite{snip} and SNIPER~\cite{sniper} on Faster R-CNN, we use the exactly the same settings, which includes multi-scale testing on [480, 800, 1400] sizes, valid ranges, and Soft-NMS~\cite{softnms}. No flipping or other enhancements are used. Our model on the same backbone achieves 47.0\% AP. It is better than the 46.1\% SNIPER.
We also compare with automatic augmentation methods, AutoAug-det~\cite{autoaug-det} and RandAug~\cite{randaug}. For a fair comparison, we train them with the same training settings to our methods on various backbones, denoted as $^\dag$. They are inferior to ours.

In addition to these common comparisons, we conduct experiments on large-scale models to push the envelope of \Ours. The baseline is the improved version of FCOS~\cite{atss} on ResNeXt-32x8d-101-DCN backbone with multi-scale training. It has 47.5\% AP in the standard single scale testing.
Without any bells and whistles, \Ours enhances this strong baseline to 48.5\% AP by + 1.0\% increase. It is further improved to 49.6\% AP on larger training images with 1200 size. When it is equipped with multi-scale testing, it is promoted to 51.4\% AP.

\subsection{Discussion}
\vspace{0.5em}
\noindent
\textbf{Understanding the searched policy.}
Tab.~\ref{tab:Searched-policy} illustrates our learned augmentation policy in details. We present each individual augmentation in the format of \{type, probability, magnitude\}. Probabilities are in the range of [0, 1.0]. The magnitude ranges for augmentations are listed in the {\em supplementary materials}. We measure it with 0 to 10 during search. This searched policy presents meaningful patterns.
\begin{itemize}
    \item Zoom-out has higher probability and magnitude than zoom-int. This matches the fact that object detectors usually have unsatisfied performance in small objects, while zoom-out benefits detecting small objects.
    \item The area ratio decreases dramatically from small scale to large scale. Note that the area ratio is searched independently in various scales from a set of discrete numbers. This phenomenon shows that augmentation involving the context (area ratio larger than 1.0) would be beneficial to small and middle object recognition.
    \item In box-level augmentations, geometric operations generally have higher probability and magnitude than color operations. It intuitively reveals that geometric operations, {\em e.g.}, rotation, translation, shearing, might have more effect than color ones in object detection.
\end{itemize}
The above patterns accord with our intuition and could provide valuable insights to human knowledge.


\vspace{0.5em}
\noindent
\textbf{Image/Feature pyramids v.s. \Ours.}
Feature pyramid network~\cite{fpn} is proposed for solving the scale variance issue in Faster R-CNN and has been widely used in this area. Here we show that our Scale-aware AutoAug could be a substitute for FPN on Faster R-CNN detector as in Tab.~\ref{tab:fasterrcnn-fpn-augs}. 
Multi-scale training is commonly known to be scale-invariant. 
However, on a clean baseline of Faster R-CNN~\cite{fasterrcnn} without FPN in 90k training iterations, it provides almost no benefits. 
In contrast, our augmentation policy improves the baseline to the performance that requires training with FPN.
Note that our augmentation policy is cost-free and requires no network modification.

\begin{table}[t]
\begin{center}
\caption{\small Scale variation issue on a clean Faster R-CNN.}
\vspace{-0.5em}
\resizebox{\linewidth}{!}{
\begin{tabular}{|l|c|ccccc|}
\hline
                 & AP   & AP$_{50}$ & AP$_{75}$ & AP$_s$  & AP$_m$  & AP$_l$  \\ \hline
ResNet-50-C4        & 34.7 & 55.7 & 37.1 & 18.2 & 38.8 & 48.3 \\
with MS train  & 34.8 & 55.6 & 37.3 & 18.9 & 39.2 & 47.6 \\
with FPN  & 36.7 & 58.4 & 39.6 & 21.1 & 39.8 & 48.1 \\
with Ours & \textbf{36.8} & 58.0 & 39.5 & 21.0 & 41.2 & 49.1 \\ 
\hline
\end{tabular}}
\vspace{-1.5em}
\label{tab:fasterrcnn-fpn-augs}
\end{center}
\end{table}

\section{Conclusion}
In this work, we present \Ours for object detection. It aims at the common scale variation issue with our search space and search metric.
Scale-aware AutoAug spends 20 GPU-days searching augmentation policies, 40 $\times$ saving compared to previous work.
Our method shows significant improvements over several strong baselines. Although the augmentation policy is searched in object detection on the COCO dataset, it is transferable to other tasks and dataset. Thus, it provides a practical solution for research and applications of augmentations in object detection.
Finally, the searched augmentation policy have meaningful patterns, which might, in return, provide valuable insights for the hand-crafted data augmentation design.

\clearpage

{\small
\bibliographystyle{ieee_fullname}
\bibliography{egbib}
}
\newpage

\twocolumn[
\begin{@twocolumnfalse}
	\section*{\centering{\em \Large Supplementary Materials \\[20pt]}}
\end{@twocolumnfalse}
]

\captionsetup[table]{labelformat={default},labelsep=period,name={Table A -}}
\captionsetup[figure]{labelformat={default},labelsep=period,name={Figure A -}}
\setcounter{table}{0}
\setcounter{figure}{0}
\appendix
\begin{figure}[t]
\begin{center}
   \includegraphics[width=0.9\linewidth]{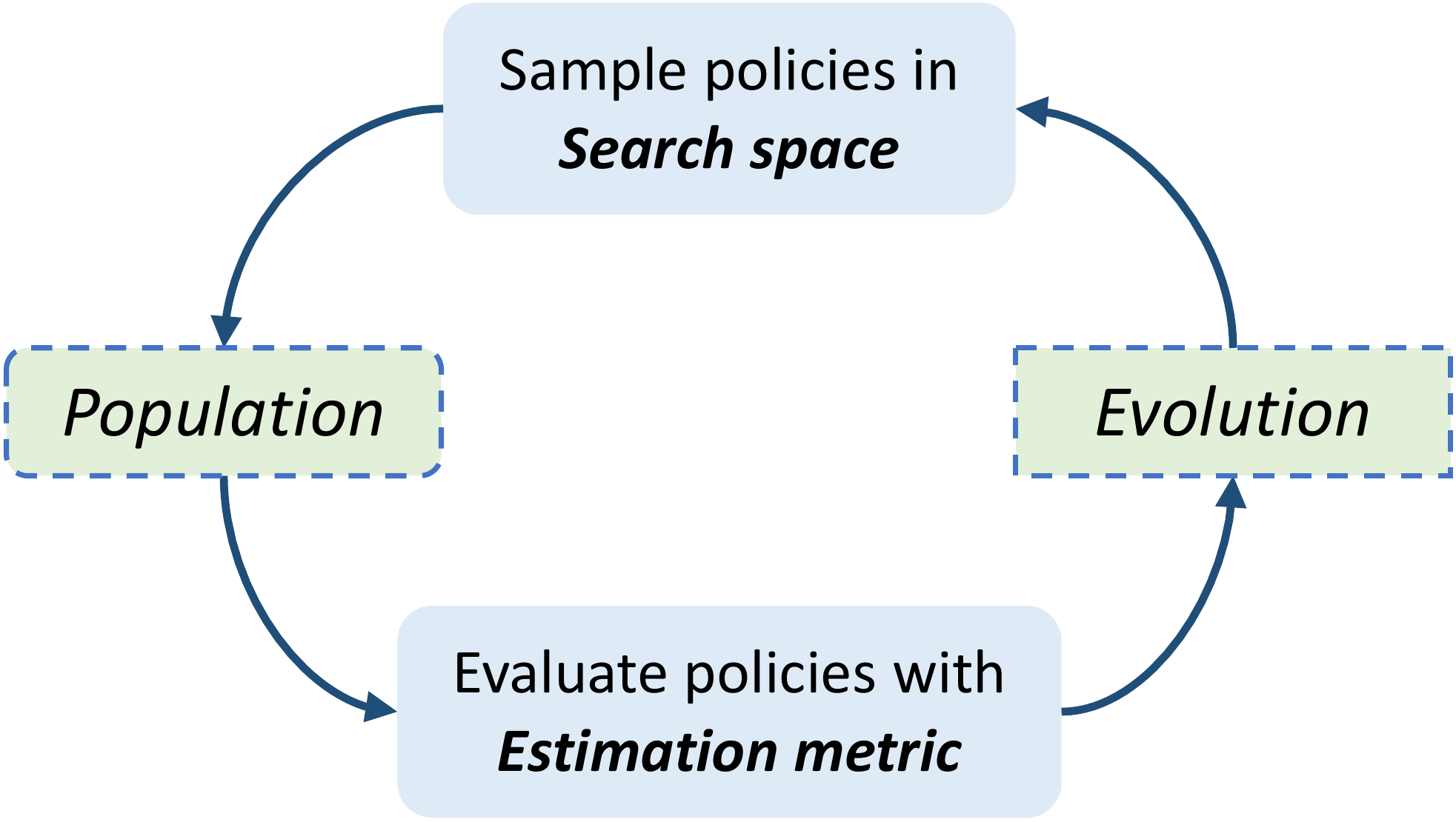}
   \caption{\small The overall evolutionary algorithm framework of our search method for learning data augmentation policies. 
   }
   \label{fig:framework}
\end{center}
\end{figure}

\section{Search framework review}
We provide a review of our overall search method framework in Fig.~A~-~\ref{fig:framework}. We adopt the evolutionary algorithm for search, where a population of data augmentation policies are randomly initialized and then evolved in iterations. During search, policies are sampled from the search space. Then, they are trained and evaluated by our estimation metric. The computed metrics serve as feedback to update. Better policies are generated in this framework over time.

\section{Derivation of Gaussian deviation}
The standard deviation of the Gaussian map can be derived as the following. Given the Gaussian map 
\begin{equation}
    f(x,y) = \exp\left(-\left(\frac{(x-x_c)^2}{2\sigma_x^2} + \frac{(y-y_c)^2}{2\sigma_y^2}\right)\right),
\end{equation}
its integration among the image can be calculated as
\begin{equation}
    V = \int_{0}^{H} \int_{0}^{W} f(x,y)\, \mathrm{d}x\mathrm{d}y \approx 2\pi\delta_x\delta_y.
\end{equation}
With the defination of the {\em area ratio} $r=V/s_{\mathrm{box}}$ and $s_{\mathrm{box}}=hw$, we can formulate their relationship as
\begin{equation}
    r = \frac{2\pi}{hw}\delta_x\delta_y. \label{eq:r-delta}
\end{equation}
Without loss of generality, the variance factors $\delta_x$ and $\delta_y$ should be correlated with the ratio of box height (width) and image height (width) to make the Gaussian map match the box aspect ratio. This can be represented as
\begin{equation}
    \delta_x/\delta_y = \left. \left(\frac{h}{H}\right) \middle/ \left(\frac{w}{W}\right) \right.. \label{eq:ratio-delta}
\end{equation}
Combining the above two equations, Eq.~(\ref{eq:r-delta}) and Eq.~(\ref{eq:ratio-delta}), we can obtain the variance factors as
\begin{equation}
    \sigma_x = h\sqrt{\frac{W/H}{2\pi}r}, \quad
    \sigma_y = w\sqrt{\frac{H/W}{2\pi}r}.
    \label{eq:sigma_sp}
\end{equation}

\begin{table}[t]
\begin{center}
\caption{\small Comparison with methods on \texttt{test-dev}.}
\resizebox{\linewidth}{!}{
\begin{tabular}{|l|cccccc|}
\hline
Method           & AP   & AP$_{50}$ & AP$_{75}$ & AP$_s$  & AP$_m$  & AP$_l$  \\ \hline 
{\em Res101:} &     &      &      &      &      &      \\
~~~PSIS~\cite{psis}          & 40.2    & 61.1 & 44.2 & 22.3 & 45.7 & 51.6 \\
~~~InstaBoost~\cite{instaboost} & 43.0    & 64.3 & 47.2 & 24.8 & 45.9 & 54.6 \\
~~~Ours           & 44.4    & 66.1 & 48.8 & 27.1 & 47.4 & 55.3 \\
\hline
{\em Res101-DCN-C4:} &   &      &      &      &      &      \\
~~~SNIP$^{\dagger}$~\cite{snip}   & 44.4    & 66.2 & 49.9 & 27.3 & 47.4 & 56.9 \\
~~~SNIPER$^{\dagger}$~\cite{sniper}    & 46.1  & 67.0 & 51.6 & 29.6 & 48.9 & 58.1 \\
~~~Ours$^{\dagger}$  & 46.9 & 68.8 & 51.7 & 30.6 & 48.1 & 58.4 \\ \hline
\end{tabular}}
\label{tab:comparison-test-dev}
\end{center}
\end{table}
\begin{table}[t]
\begin{center}
\caption{\small Comparison on larg-scale jittering.}
\resizebox{\linewidth}{!}{
\begin{tabular}{|l|cccccc|}
\hline
Method           & AP   & AP$_{50}$ & AP$_{75}$ & AP$_s$  & AP$_m$  & AP$_l$  \\ \hline 
{\em RetinaNet Res50:} &     &      &      &      &      &      \\
~~~Baseline         & 40.1    & 59.7 & 43.0 & 23.7 & 44.1 & 54.4 \\
~~~Ours  & 41.6 & 61.6 & 44.4 & 25.4 & 45.4 & 55.6 \\ \hline
\end{tabular}}
\label{tab:comparison-largescalejittering}
\end{center}
\end{table}
\begin{table*}[h]
\begin{center}
\caption{\small Details about box-level operations with their description and magnitude ranges.}
\label{tab:box-aug-details}
\resizebox{0.96\linewidth}{!}{
\begin{tabular}{|l|p{12cm}|c|}
\hline
Operation & Description   & Magnitude range     \\ \hline
Brightness  & Control the object brightness. Magnitude~=~0 represents the black, while magnitude~=~1.0 means the original. & [0.1, 1.9] \\ 
Color  & Control the color balance. Magnitude~=~0 represents a black \& white object, while magnitude~=~1.0 means the original. & [0.1, 1.9] \\ 
Contrast  & Control the contrast of the object. Magnitude~=~0 represents a gray object, while magnitude~=~1.0 means the original object. & [0.1, 1.9] \\ 
Cutout  & Randomly set a square area of pixels to be gray. Magnitude represents the side length. & [0, 60] \\ 
Equalize  & Equalize the histogram of the object area. & - \\ 
Sharpness  & Control the sharpness of the object. Magnitude~=~0 represents a blurred object, while magnitude~=~1.0 means the original object. & [0.1, 1.9] \\ 
Solarize  & Invert all pixels above a threshold value. Magnitude represents the threshold. & [0,256] \\ 
SolarizeAdd  & For pixels less than 128, add an amount to them. Magnitude represents the amount. & [0,110] \\ 
Hflip  & Flip the object horizontally. & - \\ 
Rotate  & Rotate the object to a degree. Magnitude represents the degree. & [-30,30] \\ 
ShearX/Y  & Shear the object along the horizontal or vertical axis with a magnitude. & [-0.3, 0.3] \\ 
TranslateX/Y  & Translate the object in the horizontal or vertical direction by magnitude pixels. & [-150, 150] \\ \hline
\end{tabular}
}
\end{center}
\end{table*}

\section{Augmentation operations details}
\begin{figure*}[t]
\begin{center}
   \includegraphics[width=0.93\linewidth]{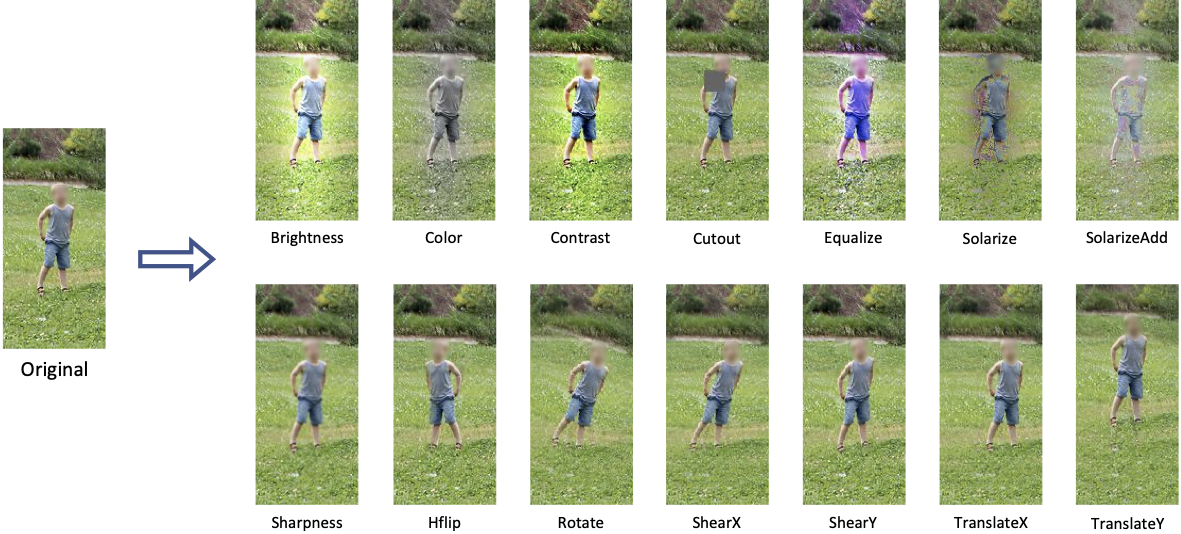}
   \caption{\small Examples on different box-level operations with magnitudes random sampled.}
   \label{fig:aug_examples}
\end{center}
\end{figure*}

We list the details about all box-level operations in Tab.~A~-~\ref{tab:box-aug-details} with their description and magnitude ranges. Besides, we provide the visualization example of these augmentations in Fig.~A~-~\ref{fig:aug_examples}.

\section{Removing context pixels}
\begin{figure*}[ht]
\begin{center}
   \includegraphics[width=0.8\linewidth]{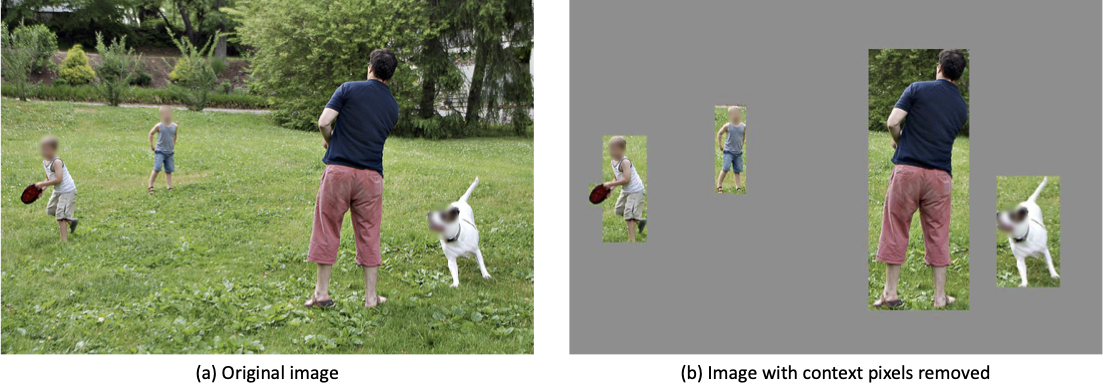}
   \caption{\small An example image of removing context.}
   \label{fig:removingcontext}
\end{center}
\vspace{-1.5em}
\end{figure*}

In Tab.~\textcolor{red}{1} in the main paper, we evaluate well-trained models on validation images whose context (background) pixels are removed. For better understanding, we provide an example image as shown in Fig.~A~-~\ref{fig:removingcontext}.

\section{Other Comparisons}
Some methods in Tab.~\ref{tab:comparison} are reported on \texttt{test-dev}. We show our counterparts on \texttt{test-dev} in Tab.~\ref{tab:comparison-test-dev}. $^{\dagger}$ denotes that the multi-scale testing technique has been used.

We also perform experiments upon the large scale jittering~\cite{spinenet}, {\em i.e.}, [0.5, 2.0] as in Tab.~\ref{tab:comparison-largescalejittering}. The baseline is enhanced to 40.1\% AP, while ours stably achieves 41.6\% AP.

\end{document}